# The Dota 2 Bot Competition

Jose M. Font and Tobias Mahlmann

**Abstract**—Multiplayer Online Battle Area (MOBA) games are a recent huge success both in the video game industry and the international eSports scene. These games encourage team coordination and cooperation, short and long-term planning, within a real-time combined action and strategy gameplay.
Artificial Intelligence and Computational Intelligence in Games research competitions offer a wide variety of challenges regarding the study and application of AI techniques to different game genres. These events are widely accepted by the AI/CI community as a sort of AI benchmarking that strongly influences many other research areas in the field.
This paper presents and describes in detail the Dota 2 Bot competition and the Dota 2 AI framework that supports it. This challenge aims to join both, MOBAs and AI/CI game competitions, inviting participants to submit AI controllers for the successful MOBA *Defense of the Ancients 2* (Dota 2) to play in 1v1 matches, which aims for fostering research on AI techniques for real-time games. The Dota 2 AI framework makes use of the actual Dota 2 game modding capabilities to enable to connect external AI controllers to actual Dota 2 game matches using the original Free-to-Play game.

**Index Terms**—Computational Intelligence, Artificial Intelligence, MOBA, Game Competition Framework

✦

## 1 INTRODUCTION

GAMES have been a recurring domain of application for computer scientists to test computer programs intended to show some sort of intelligent behavior [1]. Chess, checkers, and Go are examples of games frequently used for this task. Despite their simplicity to be simulated in a computer (two-player, turn-based, zero-sum, perfect information), these games offer a wide and complex variety of game strategies. Go has lately drawn the attention of many AI researchers, and Google's AlphaGo AI agent recently succeeded at beating the European human Go champion [2].

Video games used as AI benchmarks offer interfaces to which external AI modules can be connected, with the aim of evaluating the performance of those modules while solving particular tasks within the game domain [3]. These benchmarks are usually involved in Internet-based competitions, some of those organized during international research conferences. Games as AI benchmarks are the most influential AI/CI area, being strongly influential to other areas such as General Game AI and NPC Behavior Learning. Though each benchmark generally presents a single problem, together they compose an extensive set of specific problems to be addressed by researchers, which composes a good framework for continuous optimization in AI and CI. A notorious exception to this is the General Video Game AI Framework and its related competitions [4], which involve creating AI that can play a wide range of classic games included in the platform, all of them implemented using the Video Game Description Language [5].

A very popular form of competition in video games, though not related to research, are eSports: organized international multiplayer competitions that involve professional game players. Initially very popular in Asian markets, the worldwide eSports market is valued in $1.5B in 2017, with an estimated revenue growth of 26% by 2020 [6]. The video game genres most commonly played in eSports are real-time strategy (RTS), fighting, first-person shooters, and MOBA. This stands for Multiplayer Online Battle Arena [7], a sub-genre of RTS in which a player controls a single character (often called a hero) in a team who competes versus another team of players to destroy each other's base, following a zonal or strategic point control. Characters' abilities are improved over time, being equipped and/or leveled up following a progression commonly found in role playing games (RPG). Maps are commonly symmetrical and periodically spawn computer-controlled units that support the competing teams. Heroes have various unique abilities that are be improved over time, which contributes to the team's game strategy. Some MOBAs also introduce economic mechanics such as gold and items. MOBAs are played real-time and typically in isometric perspective, providing both collaborative and competitive play experiences. MOBAs currently top the ranking of the most played and profitable Massive Multiplayer Online games, with the successful League of Legends (LOL) earning more than $150M monthly [8]. Another very popular MOBA, Defense of the Ancients 2 (only referred to as *Dota 2*), hosts the biggest prize pools in eSports, which was higher than $24M in the 2017 competition [9]. Moreover, the estimated lifetime *Dota 2* prize pool is currently higher than $135M [10].

This paper presents the Dota2 2 bot competition, an AI/CI competition about writing controllers for the heroes taking part in Dota 2 matches. This is the only AI/CI competition that uses a MOBA as the framework for evaluating the performance of AI controllers. Due to the collaborative and competitive nature of MOBAs, this competition focuses in the development of AI agents that communicate and collaborate between them, using no tools but the ones included in the game itself for team communication, i.e. map pings

- *Jose M. Font is with the Department of Computer Science and Media Technology, Malmö University, Sweden.*
  *E-mail: jose.font@mah.se*
- *Tobias Mahlmann is with the Department of Cognitive Science, Lund University, Sweden.*
  *E-mail: t.mahlmann@gmail.com*

and pre-defined phrases. Section 2 provides a deeper insight into existing game frameworks and competitions, as well as previous research regarding Dota 2. Section 3 describes in detail the game with its core mechanics, as well as the specifics of the Dota 2 bot competition and the Dota 2 AI framework that was developed to make it possible.

## 2 RELATED WORK

Different competitions address various AI/CI problems through different game frameworks, being IEEE's Conference on Computational Intelligence in Games one the events in which most of them take place on a yearly basis [11]. The Ms Pacman competitions [12] have been running for more than 10 years now, asking participants to write controllers for both Pacman and the ghosts using the Ms Pacman framework (an open implementation of the sequel to the original Pacman). The Geometry Friends competitions [13] asks for controllers that must succeed at playing levels in a custom cooperative video game. The Starcraft AI competition has also been running for long and under different formats (CIG, AIIDE, SSCAI) [14]. Thanks to the BWAPI and some other external tools, it makes possible to load custom AI controllers for the original Starcraft: Brood War game. The microRTS competition [15] pursues similar goals but replacing an actual RTS game by a custom simplified version that keeps the research aspects of the competition from been relegated to a second plane due to common engineering problems when dealing with commercial RTS games. The Fighting Game competition [16] follows a similar approach but switching the game genre to a Street Fighter-like fighting arcade named FightingICE. The latest version of this framework also allows the creation of visual-based AI controllers, similar to the main goal of the Visual Doom AI competition [17], that uses the ViZDoom platform to build AI controllers that use the screen buffer as the only input. Other controller oriented competitions are the Text-Based Adventure AI competition and the Showdown AI Competition, dedicated to bots for classic text adventure games and Pokemon-like battles, respectively.

The aforementioned address the problem of AI controller generation from different perspectives and game genres, but this is not the only existing topic in AI/CI competitions. The previously mentioned General Video Game AI competition [4] presents different tracks, being one of them dedicated to AI controllers but the other two to level generation and game rules generation. Similarly, the Angry Birds AI competition [18] hosts two tracks related to the Angry Birds game: one dedicated to controller generation and the other one to level generation. Finally, the Game Data Mining competition [19] focuses in game analytics and player behavior understanding by applying machine learning techniques to existing recorded gameplay from NCSoft's Blade and Soul.

All these competitions and frameworks promote research in many different techniques, such as Monte Carlo tree search, evolutionary computation, neural networks, and machine learning, as well as explore their application to various game genres, like RTS, FPS, fighting, adventure, and puzzle. Despite their popularity, not much research has been carried out regarding AI and MOBAs [20]. Open AI created a bot which beats the world's top Dota 2 players at 1v1 matches [21], though the specifics of its learning process have not been made public, and there is little known apart from that it bases upon recorded experts' gameplay. Drachen et al. [22] presented novel measures for data-driven player behavior analysis on Dota 2, as well as a method for obtaining and visualizing accurate spatio-temporal data from the game, highlighting its importance to aid players in visualizing, analyzing, and improving their performance [23]. Through the Bot of Legends API (similar to BWAPI for Starcraft), Silva and Chaimowicz [24] developed an AI controller based on influence maps for its most popular competitor, League of Legends.

## 3 THE DOTA 2 BOT COMPETITION

Motivated by the raise of MOBAs among the general audience plus the existing research on the field, showing the potential of MOBA games to act as frameworks for AI benchmarks, in this section we present the Dota 2 bot competition and the Dota 2 AI framework that was developed to make it possible.

### 3.1 The game: Defense of the Ancients 2

Defense of the Ancients 2 (Dota 2) is a Free-to-Play (F2P) MOBA, originated as a sequel to the community-created Defense of the Ancients (Dota) mod for Blizzard Entertainment's *Warcraft III: Reign of Chaos*, released in 2003. The first version of the Dota was created as an entire new scenario for Warcraft III using its world editor, inspired by a popular StarCraft map called "Aeon of Strife". After several improved versions, the mod became so popular within the e-sports scene that Valve started developing Dota 2 as a standalone game. Dota 2 stands today as one of the most played games on Steam (http://steamspy.com/app/570).

Two teams (named Radiant and Dire) of five players play against each other in short matches (from 30 to 60 minutes on average), with each team defending their own separate base, located both on the opposite corners of a symmetric square map. Each player controls a hero chosen from the 118 available, designed with unique abilities and skills. Following a typical structure in RPG games, heroes are equipped with items purchased from shops with gold, which can by killing enemy heroes and monsters; and their abilities and skills are upgraded using experience points. Heroes respawn in their team's base some time after being killed an unlimited number of times. The respawn time is longer the higher the hero's level. All heroes start from their base level after every match. Though the game is intended to be played online by human players, built-in AI bots can control some of the heroes participating in the matches as stand-ins, which has turned into a recommended practice for beginners while learning the game mechanics and components.

One of most representative components of any MOBA is the map (Figure 1), which is divided in two halves owned by each of the teams. Each team's base features an *ancient*, the main building. A team wins by being the first one in destroying the opponent's ancient. There are three main pathways (lanes) that connect both ancients through the map, named as *top*, *middle*, and *bottom*. The ancients are guarded by a series of defensive towers symmetrically distributed along these pathways. Both bases regularly spawn computer-controlled monsters, called creeps, that automatically attack the opponent's buildings, creeps, and heroes. There are a variety of smaller paths and zones between the main pathways, opening the gameplay to different tactics and strategies. These areas are commonly referred to as *the jungle*.

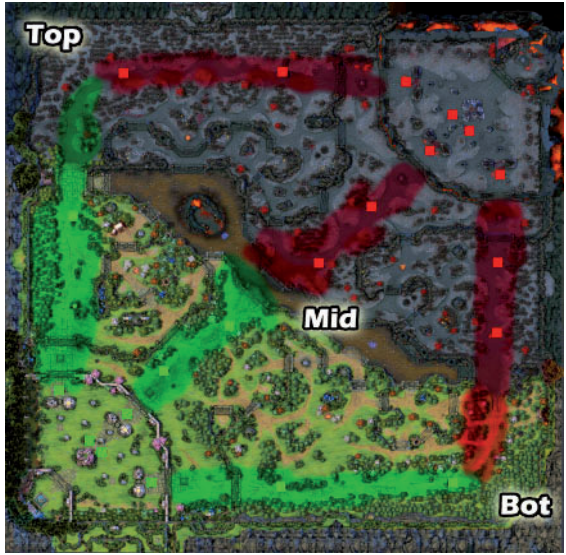

Fig. 1. The Dota 2 map as of game version 7.00. Radiant (green) and Dire (red) bases are in opposing corners (bottom left and top right, respectively), connected by the three main lanes (commonly named as Top, Mid, and Bot). All towers are marked by the colored squares.

### 3.2 The competition

The Dota 2 bot competition invites participants to write AI controllers (bots) for heroes in Dota 2 matches. Bots participate in a series of 1v1 solo mid matches [25] facing subsequently different heroes controlled by the game's built-in AI. Submitted controllers are ranked by the number of matches won. The 1v1 solo mid is one of the official Dota 2 game modes, implying the following special conditions:

- Only one hero per team participates.
- Both players can choose the same hero, and the opponent's choice is not revealed until theirs is made.
- Creeps only spawn in the middle lane.
- Each team starts with a free courier (non-attacking NPC that transports purchased items between heroes and shops).
- The match ends when the first tower (T1) in the mid lane is destroyed, with the tower owner losing the match. Achieving two kills has been removed as an end condition to make it more challenging, due to some bots dying frequently on very early stages.

The ultimate goal for the competition is to develop AI controllers that perform well while they coordinate and collaborate with their teammates, with no other communication channels besides the chat wheel and map pings. Nevertheless, the current version of the Dota 2 AI framework only supports one external controller per match. 5v5 matches can be hosted with only one hero played by the external controller and the remaining 9 heroes managed by the built-in AI. This AI coordinates its bots but does not handle any incoming communication to-from the external bot. Since Dota 2 does not offer the possibility to speed up the gameplay, all matches have to be played in strict real time.

### 3.3 The Dota 2 AI framework

The Dota 2 bot competition is made possible thanks to the Dota 2 AI framework by Tobias Mahlmann, a Dota mod developed ad-hoc for the competition. Dota 2 offers users a LUA sandbox engine for them to script and mod various aspects of the game [26]. This can be used by people who would like to create new content for the game such as game modes, heroes, and abilities.

The Dota 2 AI framework makes use of this possibility to create a copy of the basic game mode and the default map, including an additional module that acts as a proxy between the game and a web service through JSON objects. Though this architecture is not very straightforward, it is unfortunately the only way to establish a communication between the game and an external process, i.e. an external AI controller. The complete source code and its documentation can be found in [27], being also available from Steam's Workshop website [28].

Notice that, while it would eventually be possible to write and run a bot directly in the LUA sandbox, this option has severe disadvantages. It would prevent re-using existing libraries coded and tested in other languages. It would also keep bots from having any learning capabilities or any other time consuming algorithms that need an asynchronous execution, since everything runs in the same process as the main game. Last but not least, coding bots in the LUA sandbox would require additional measures to prevent authors from cheating.

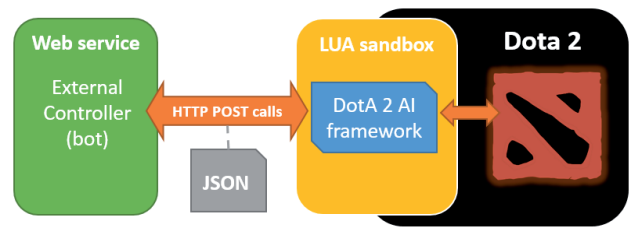

Fig. 2. Overview of the Dota 2 AI framework and its integration with the Dota 2 game. The AI framework extracts the information from the game state and makes it available to external controllers via JSON formatted files transferred via HTTP POST requests.

#### 3.3.1 Setting up the framework

Figure 2 shows the architecture and the overall workflow of the Dota 2 AI framework, the game, and the external AI controller. The main components are the Dota 2 game, the LUA mod, and the external controller. Since the framework is formally a game mod, the following steps are required to set it up:

1) Download and install a copy of Dota 2 from a Steam client, together with the Dota 2 Workshop

    Tools DLC. Both items are available for free from Steam (http://store.steampowered.com/app/570/Dota_2/).
2) Download the Dota 2 AI framework and copy the files from the Dota 2 AI Addon folder into the Dota 2 game installation folder.
3) Launch the Dota 2 mod by loading the dota2ai project from the Dota 2 Workshop Tools.

These steps will prepare and run a Dota 2 mod match on the standard map. The mod's LUA scripts will access the game during the whole match, making many game state features available for the external web services at every game tick in real time.

### 3.3.2 Designing and connecting the external controller

Participants must code their own controller (bot) in any given language and framework that runs it as a web service. Running the web service after the former set up will connect the bot to the Dota 2 match, interacting with it in both directions. This communication is achieved by HTTP POST calls, and all the information regarding game states and actions is transferred in JSON formatted files.

The Dota 2 AI Addon folder contains all the LUA scripts implementing the mod, including a *config.lua* file in which the base URL where the web service is listening from must be specified (i.e. http://localhost:8080/Dota2AIService/). The framework's documentation [27] lists all the functions in the LUA scripts that can be used for bot-game communication. For every LUA function listed, the web service must provide an homonymous listening function, whose URL must also be specified in the *config.lua* file. I.e. the LUA function */test* must be paired with a listening function in the web service whose URL is http://localhost:8080/Dota2AIService/test. These functions must accept POST requests with the MIME type *application/json*. This is the data type for receiving and sending information from and to the game mod.

### 3.3.3 Formatting the messages

The LUA functions serve two main purposes: notifying of key events during the gameplay (some of them in need of a response from the controller), and transferring the relevant information about the game state to the controller.

The *Select* function is a good example of a **key event**. Select is called by the game right before the game starts. The game expects the player (the controller) to determine which hero the game should spawn. To answer the *Select* call, the controller must provide a function under the URL http://localhost:8080/Dota2AIService/select, that handles messages with this JSON format:

```
Function: Select

Input parameters: none

Return values:
    String hero
    String command
```

```
Return message example:
    {
        "hero":"npc_dota_hero_lina",
        "command":"SELECT"
    }
```

The game mode is currently set to *all pick*, so all the heroes in the game are available. No input parameters are required for hero selection, but the game is expecting two returned values: the id of the chosen hero and the *SELECT* command. All hero ids are listed in the Dota 2 scripting specification [26]. The sample returned message indicates that the controller has selected the hero named *Lina*. It is important to notice that the function calls are synchronized with the game, so the game waits until the external controller replies. As we assume that writing a controller for Dota 2 is already challenging enough, we do not impose timing restrictions. This might change in the future. In the current version a warning is simply sent to the game's console.

Another good example is the *Chat* function, called whenever another player submits a text message in the game. The message is simply transmitted to the bot, which may or may not implement a reaction to it. The specification for this function is the following:

```
Function: Chat

Input parameters:

    Boolean teamOnly
    String text
    Integer player

Return values: none

Input message example:
    {
        "teamOnly":false,
        "text":"Humans are n00bs!",
        "player":5
    }
```

This function returns nothing but has three input parameters: *teamOnly*, that equals true when the message is only for the team chat, the *text* with the message content, and the *player* id of the sender.

The **current game state** is transferred to the controller by the *Update* function. This is called every time the game's internal state is updated (approx. every 33 milliseconds). The only input parameter is *Entities*, the complete game state as perceived in-game by the bot-controlled hero. *Entities* is a JSON array that contains a JSON object (entity) per each relevant game state component. Each entity has an internal id and a class, describing the game object the entity is related to. The list of classes describing all game objects is included in the Dota 2 scripting specification [26]. The following is an example of a JSON object depicting an enemy tower, including several features like health points, level, attack range, etc.

```
"760":
{
    "level": 1,
    "mana": 0,
    "disarmed": false,
    "health": 1300,
    "alive": true,
    "attackRange": 700,
    "team": 3,
    "blind": false,
    "dominated": false,
    "origin":
      [
        -4736,
        6016,
        383.99987792969
      ],
    "type": "Tower",
    "rooted": false,
    "name": "dota_badguys_tower1_top",
    "deniable": false
}
```

Finally, the LUA framework is ready to receive, process, and apply the **returned values** by the external bot. A returned value is the command the hero will apply in the next game tick. Like playing the game as a human, the hero will execute the last given command until something stops him or a new command is given.

When no new commands have to be returned by the bot, it should reply with a *NOOP* (no operation) command. Otherwise, the list of available commands, related to the hero's available in-game actions, is: *Move, Attack, Cast, Buy, Sell,* and *Use_Item*. The following are three sample return values containing the *Move*, *Attack*, and *Cast* actions, respectively.

```
{
    "x" : "4000",
    "y" : "4000",
    "z" : "380",
    "command" : "MOVE"
}

{
    "target" : 370,
    "command" : "ATTACK"
}

{
    "ability" : 3,
    "target" : 370,
    "command" : "CAST"
}
```

*Move* drives the hero towards the specified *x, y, z* coordinates. *Attack* makes the hero attack the *target* (specified by its id), and *Cast* triggers the hero's specified *ability* on the chosen *target*. The Dota 2 AI framework documentation explains every action in detail [27].

### 3.4 The Java example bot

The above describes all that is required to create an external controller that runs on the Dota 2 AI framework, therefore a suitable bot for participating in the Dota 2 Bot Competition. Additionally, the Dota 2 AI framework provides a sample ready-to-use bot for the participants to learn the basics of the platform. This integrated solution is distributed with the framework [27] under the *Dota2BotFramework* and the *Example Bot* Java projects.

The *Dota2BotFramework* is a web application that runs on *NanoHTTP*, providing a web service and a basic abstraction model for a bot to use. This implements a Java interface between the *Example Bot* and the LUA scripts, handling all the calls, returned values, and message formatting explained in the previous subsections. The *Example Bot* is a simple state machine for a Lina hero, that can go to the middle lane, attack creeps and enemy heroes, and randomly cast spells. She also retreats if her health falls under a certain percentage, as well as responds to basic commands sent to team chat like *lina go* or *lina buy tango* (a tango is a frequently used item in Dota 2).

## 4 CONCLUSIONS AND FUTURE WORK

Here we have presented the first version of the Dota 2 Bot competition, together with the Dota 2 AI framework that makes possible for an external bot to take control of a hero during standard Dota 2 matches. This work is motivated by the popularity of MOBA games in the eSports scene, together with the current lack of scientific competitions in this genre and the interesting potential that it offers to work on AI controllers for real-time games with a special focus on team communication and cooperation.

The framework makes use of the Dota 2 LUA sandbox to create a mirrored version of the game in which a communication channel is open to an external web service during the gameplay. This allows to code external controllers as web services with complete interaction capabilities with the game. In the current version of the framework, the Dota 2 Bot competition consists of a series of 1v1 solo mid matches, in which an external bot faces different Dota 2 heroes until the first tower in the mid lane is destroyed. The bots are ranked according to their number of victories. Participants can develop bots for each of the 118 available characters in the game, which allows them to create AIs that exploit their specific abilities and skills.

The framework provides a ready-to-use Java example bot for participants to learn the principles of bot development for the platform, as well as a more open approach in which participants can code their bot in their preferred language with the sole restriction of encapsulating it as a web service.

The Dota 2 Bot competition was already part of the competitions in IEEE's CIG 2017, though no participants submitted solutions on that occasion. Authors intend to submit this competition to the upcoming relevant events in the field, inviting researchers to submit their controllers with the aims of fostering research on AI techniques suitable for this popular game genre. These initial submissions will also be used to improve the competition, as well as the framework and its documentation, to offer a better experience in the subsequent iterations.

As for now the framework offers a single track competition on 1v1 matches, the next improvement will be to explore the possibility of updating the mod to enable the connection of more than one external bot at a time. This would support the addition of different tracks for the competition, including 5v5 matches with actual bot communication, opening scope for strategies based on team cooperation using the in-game chat and real-time gamestate analysis.

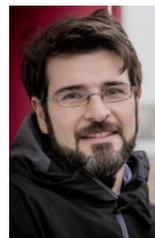
**Jose M. Font** was born in Jerez de la Frontera, Spain, in 1983. He received a B.Sc. degree in Computer Science from the Universidad Antonio de Nebrija in 2006, and the M.Sc. and the Ph.D. in Artificial Intelligence from the Universidad Politécnica de Madrid in 2009 and 2012, respectively. He is a tenured assistant professor at Malmö University since 2016. His research focuses on artificial intelligence and computational intelligence in games, procedural content generation, and mixed-initiative creative tools. He is also active in gamification, e-learning, and purposeful games. He is also the general chair of the ACM's Foundations of Digital Games 2018 conference.

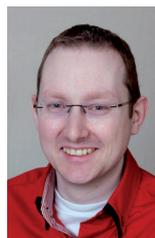
**Tobias Mahlmann** was born in 1982 in Hannover, Germany. He holds the degree of Diplom-Informatiker from the Technical University of Braunschweig, Germany and further a PhD from the IT University of Copenhagen, Denmark. His focus areas in computer science are Artificial Intelligence and Video Games. He is an active researcher on general AI for playing games, procedural content generation, automatic game evaluation, and player behavior prediction. He is currently working on Virtual Reality applications as the technical director at Moon Mill Studios located in Copenhagen, Denmark.